\documentclass{article}

\usepackage{microtype}
\usepackage{graphicx}
\usepackage{subfigure}
\usepackage{booktabs} 
\usepackage{enumitem}

\usepackage{hyperref}

\usepackage[accepted]{icml2025}

\usepackage{amsmath}
\usepackage{amssymb}
\usepackage{mathtools}
\usepackage{amsthm}

\usepackage[capitalize,noabbrev]{cleveref}

\theoremstyle{plain}

\theoremstyle{definition}

\theoremstyle{remark}

\usepackage[textsize=tiny]{todonotes}

\icmltitlerunning{Interpretable Diffusion Models with B-cos Networks}

\begin{document}

\twocolumn[
\icmltitle{Interpretable Diffusion Models with B-cos Networks}



\icmlsetsymbol{equal}{*}

\begin{icmlauthorlist}
\icmlauthor{Nicola Bernold}{infk}
\icmlauthor{Moritz Vandenhirtz}{infk}
\icmlauthor{Alice Bizeul}{infk}
\icmlauthor{Julia E. Vogt}{infk}

\end{icmlauthorlist}

\icmlaffiliation{infk}{Department of Computer Science, ETH Zürich, Switzerland}

\icmlcorrespondingauthor{Moritz Vandenhirtz}{moritz.vandenhirtz@inf.ethz.ch}

\icmlkeywords{Explainable AI, Interpretability in Machine Learning, B-cos Networks, Diffusion Models, Generative Models, ICML}

\vskip 0.3in
]



\printAffiliationsAndNotice{}  

\begin{abstract}
Text-to-image diffusion models generate images by iteratively denoising random noise, conditioned on a prompt. While these models have enabled impressive progress in image generation, they often fail to accurately reflect \textit{all} semantic information described in the prompt---failures that are difficult to detect automatically. In this work, we introduce a diffusion model architecture built with B-cos modules that offers inherent interpretability. Our approach provides insight into how individual prompt tokens affect the generated image by producing explanations that highlight the pixel regions influenced by each token. We demonstrate that B-cos diffusion models can produce high-quality images while providing meaningful insights into prompt-image alignment.
\end{abstract}

\section{Introduction}
Text-to-image generative models have achieved impressive results, yet, interpreting their decision process remains a significant challenge, making prompt adjustments a tedious manual process~\cite{xu2024prompt}. Various post-hoc explanation techniques, such as LIME~\cite{ribeiro2016should}, SHAP~\cite{lundberg2017unified}, or LRP~\cite{bach2015pixel}, have been proposed to better understand deep neural networks. However, these approaches often fall short of faithfully representing the underlying behavior of the models they aim to explain \cite{bohle2022b}.
Hence, inherently interpretable deep neural networks have attracted growing interest \cite{bohle2024b,rio2024suitability}. Unlike post-hoc explanation methods, these models are designed with built-in architectural constraints that enable the model to produce a faithful explanation for its output.

Integrating interpretable modules into existing architectures can be challenging, often requiring tedious adjustments or being infeasible due to the constraints these modules impose. To address this, \citet{bohle2022b} introduce B-cos networks, which serve as drop-in replacements for standard neural network components. This design allows interpretable elements to be seamlessly incorporated into existing architectures with minimal impact on predictive performance, while providing faithful visualizations of input-output relationships and even neuron-level explanations.
So far, B-cos networks have primarily been developed for discriminative tasks in computer vision and natural language processing~\cite{bohle2024b}. 

In this work, we extend B-cos networks to text-to-image diffusion models. While diffusion models \cite{ho2020denoising,song2021denoising} have emerged as the state of the art for generating images, they remain difficult to interpret and evaluate. In particular, assessing how accurately the semantic concepts expressed in the prompt are reflected in the generated images is a key challenge—one that hinders the evaluation of image-text alignment. Our work seeks to address these challenges through the following contributions:

\begin{enumerate}[itemsep=0pt, topsep=0pt]
    \item We identify key considerations and challenges of integrating B-cos networks into text-to-image diffusion models, which provides a foundation for future work.
    \item We integrate B-cos networks into Stable Diffusion and demonstrate its generative capabilities as well as faithful attribution of relevant input tokens.
    \item Leveraging its inherent interpretability, we show that the proposed method can identify failure generations where the model's outputs do not align with its prompt.
\end{enumerate}

\section{Background}
Neural networks can be locally approximated by an affine transformation of the form \( f(x) = W(x)x + b(x) \), where \( W(x) \) and \( b(x) \) are input-dependent tensors that provide a faithful summary of the model's behavior at a given point~\cite{srinivas2019full}.
To transform this observation into a basis for model explanations, B-cos networks \cite{bohle2024b} aim to eliminate the bias term \( b(x) \), ensuring that \( W(x) \) alone encapsulates the full information of the model's computations. For an image classifier logit, $W(x)$ has the shape of an image and thus can directly be visualized. However, such visualizations typically only display local features, which do not correspond to the global, human-understandable concepts in the image. To address this, B-cos networks employ the so-called B-cos transformation in place of the traditional neurons: 
\begin{equation}
    \begin{split}
        f_\text{B-cos}(x; w) &= \left(\lvert \text{cos}(x,w) \rvert^{B-1} \times \hat w\right)^Tx \\&= w^T(x)x
    \end{split}
\end{equation}
with hyperparameter $B$, the cosine similarity $\text{cos}(x,w)$ between $x$ and $w$, and $\hat w\!=\!\frac{w}{||w||_2}$, $w$ scaled to unit-norm.   

Such B-cos transformations are thus accurately summarised by a single linear transformation $W(x) \text{ s.t. } f(x) = W(x)x$. As there is no implicit or explicit bias term, this summary is complete and for $B>1$, the cosine similarity puts additional pressure on the input-weight alignment which has been shown to make the explanations more human interpretable and lead to a better localization of the task-relevant input features. These properties also extend to layers and stacks of such transformations which implies that arbitrarily complex B-cos networks can be summarised by a single dynamic linear transformation. 
The input-weight alignment property is the key concept of B-cos networks:
\begin{equation}
    B \gg 1 \land \lvert f_\text{B-cos}(x;w) \lvert \gg 0 \implies \lvert cos(x,w) \rvert \approx 1
\end{equation}
This means that if the output of the B-cos transform is non-negligible, the weights are aligned with the input. Or in other words, the explanation, which is based on the scaled weights, corresponds roughly to the relevant input for a particular output neuron. 

\section{B-cos Diffusion Models}
\label{sec:method}

To leverage the interpretable qualities of B-cos networks for diffusion models, the neural network backbone needs to be replaced by a B-cos network. Ideally, this should be a drop-in procedure such that it can be used together with any architectures and techniques that have been shown to work well for conventional diffusion models. In the following sections we discuss further considerations, challenges, and solutions for diffusion models in general, but also focus on text-to-image Stable Diffusion 2.1 \cite{Rombach_2022_CVPR} for more specific details. For a more thorough discussion of related work, we refer to \Cref{appendix:rel_work}.

\subsection{Inputs and Biases}

Neural networks themselves often contain learnable or constant biases, which are removed when transforming a neural network architecture into a B-cos network. However, when generating images with a conditional diffusion model, the neural network backbone is applied multiple times and various operations are performed between the individual inference runs to turn the predictions into the sample at the next timestep. These computations outside of the neural networks still introduce biases and the end-to-end B-cos summary thus is no longer complete. Specifically, given a noisy image $x_t$ at timestep $t$ with conditioning $c$, the U-Net predicts some quantity which is used to compute the predicted mean $\hat \mu(x_t,c,t;\theta)$, such that $x_{t-1}$ can be sampled from some distribution:
\begin{equation}
    x_{t-1} \sim \mathcal{N}(\hat \mu(x_t,c,t;\theta), \sigma_t I)
\end{equation}
The noise introduced through this step is a bias which cannot be explained with respect to any of the inputs. However, applying deterministic sampling schemes, such as e.g. DDIM \cite{song2021denoising} with $\sigma_t = 0$, allow us to fully avoid introducing new biases by sampling from a distribution between the U-Net inferences.

Moreover, the diffusion model backbone has multiple inputs. For example a text-to-image diffusion model backbone receives a noisy image, an encoded text prompt and a timestep encoding as input. B-cos explanations are designed to explain how a single input affects the output. Thus, the multiple inputs practically behave like biases towards each other. Omitting the noisy image or the conditioning as input cannot be done without fundamentally modifying the architecture such that it is no longer a conditional diffusion model. However, previous work has shown empirically that the inclusion of a positional encoding still leads to good explanations despite introducing a bias.


\subsection{Cross-Attention}
Cross-attention is central to conditional diffusion models. Given a noisy image input $X$ and conditioning $Y$ (e.g., a text prompt), the computation is:
\begin{equation}
    \begin{split}
        \text{Cross-Att}&(X,Y;Q,K,V) \\
        &= \text{softmax}\left( XQK^TY^T/ \sqrt{d_k}\right)YV\\
        &= \text{A}\left( X,Y\right)YV
    \end{split}
\end{equation}
Similar to the dynamic linear view of self-attention in \citet{bohle2024b}, both the attention scores $A(X,Y)$ and the value transformation $V$ can be seen as dynamic linear mappings from $Y$ to the output. To align conditioning with the weights, we replace the linear value projection $YV$ with a B-cos layer, making the full cross-attention a B-cos transformation modulated by attention scores. This preserves the alignment property and can be expressed as a single dynamic linear matrix $W(Y)$, where we treat $X$ as a non-additive constant.
As our focus is on explaining outputs with respect to $Y$, the module summary of interest is complete.

\subsection{Input and Target Encoding}
\label{subsec:encoding}
The backbone of a diffusion model takes an image-shaped input and produces an output of the same shape. Conventionally, the image is encoded using three channels $(r,g,b)$. In order to have the input-alignment guarantee, we need non-negligible outputs. Moreover, the output magnitude of B-cos networks is bounded by the input magnitude. Also, with the aforementioned encoding, the cosine similarity cannot distinguish between dark and bright colors. Thus, we propose to use the following encoding for input and target, which is the same encoding as used for the inputs of B-cos vision classifiers:
\begin{equation}
    \text{Enc}(r,g,b) = (r,g,b,1-r,1-g,1-b)
\end{equation}
As the output of the backbone is fed into the input again and as we want to avoid introducing new biases, having the same input and target encoding is the obvious choice. However, such an encoding only works if $x_{t-1}$ or $x_0$ is predicted directly. Many networks are trained to predict $\epsilon_t$ -- the noise that was added to the image $x_0$ to obtain the noisy sample $x_t$. To encode the noise in the above way, we employ a diffusion process with a noise mean and variance of $0.5$. The details can be found in \cref{appendix:diffusing}. We have experimented with predicting the noise but empirically found that predicting $x_0$ is the only option with promising denoising capabilities. 

The relationship between the prompt and the generated image is of particular interest. 
Stable Diffusion conditions the U-Net on the token embeddings of a CLIP encoder. Because of cross-attention in CLIP, these tokens may capture contextual information beyond their associated subword units. 
To alleviate this issue, we use the subword embeddings of CLIP directly and mask all padding tokens, including the \emph{SOS} and \emph{EOS} tokens. This way, only tokens with a semantic meaning can contribute to the output and the explanations do not need to be disentangled.

Similarly, Stable Diffusion also employs a variational autoencoder (VAE) to project the images onto a latent space in which the denoising is performed. Instead of separately training a B-cos VAE, we omit the VAE and directly perform pixel-space diffusion on a lower $64\!\times\!64$ resolution. Our setup thus focuses on integrating B-cos networks into a diffusion model's backbone: the model's UNet.

\subsection{Interpretability and Explanations}

Given an encoded prompt x and the model summary W(x) of some sampling run, one can use that model summary to reconstruct the model output together with the prompt x:
\begin{equation}
    R(x) = R_{rgb}(x) \mid\mid R_{1-rgb}(x) = W(x)x
    \label{eq:reconst}
\end{equation}
This reconstruction $R(x) \in \mathbb{R}^{H\times W\times C}$ has $C=6$ channels. As the diffusion models contain biases, the summary does not capture all model computations and hence $R(x)$ is not equal to the original model output. However, the six channels contain redundant information. Hence, the model output and the reconstruction can be split into the two images with the three channels corresponding to $(r,g,b)$ and $(1-r, 1-g,1-b)$, respectively. The sum of each channel in $R_{rgb}$ and $R_{1-rgb}$ should thus add up to one. To enforce this and potentially recover information lost due to biases, the normalized reconstruction can be computed as follows:
\begin{equation}
    R_{\text{normalized}}(x) = R_{rgb}(x)/(R_{rgb}(x)+R_{1-rgb}(x))
    \label{eq:reconst_norm}
\end{equation}
These reconstructions allow us to gain insights into the information loss due to the biases. The difference between the normalized reconstructions and the model output being small implies that the model summary captures the relevant model computations.  

To compute the attribution of the i-th prompt token of a generated sample, we use its embedding $x_i$ and the i-th row in $W(x)$ to compute the contributions to each pixel and channels, then aggregate those and normalize the score across all tokens. We refer to this measurement as the \textbf{relevance score}:
\begin{equation}
    S_i(x) = \frac{\lvert\sum_{h,w,c} W(x)_i x_i\rvert}{\sum_j\lvert\sum_{h,w,c} W(x)_j x_j\rvert}
    \label{eq:relevance}
\end{equation}
Computing $W(x)$ is extremely computationally expensive. Storing it during the forward pass is not trivially possible without obtaining prohibitively large tensors. Thus, backward passes are typically used to compute the summary for the interesting outputs. In our case, that's not a single class logit but a full image with six channels. However, to compute the attribution score, $W(x)$ does not need to be fully materialized. The summation over the pixels and channels can be fused into its computation. That is, the attributions for all tokens can be computed in a single backward pass. 

\section{Experiments}
\subsection{Experimental Setup} 
For our experiments, we modified Stable Diffusion 2.1 \cite{Rombach_2022_CVPR} to use a B-cos U-Net with the embeddings of the pretrained CLIP model as described in \cref{subsec:encoding}. The models were trained for one million steps on a subset of LAION-2B-en-aesthetics \cite{schuhmann2022laion} focusing on five selected objects: banana, cat, goat, flamingo and penguin. We included 20'000 image-caption pairs for each object. We refer to our model as \emph{B-cos $x_0$}. Further details are specified in \cref{sec:model_details,sec:training}.

As baseline we also train \emph{Vanilla} Stable Diffusion with the same settings. We also include \emph{B-cos Clip eps} and \emph{B-cos Clip $x_0$}, which use the frozen Clip Encoder of Stable Diffusion 2.1 and are trained to predict the reparameterized zero-mean unit-variance noise and the target $x_0$, respectively, to provide a comparison of the sample quality. For a quantitative comparison, we compute the FID score~\cite{heusel2017gans} using 100'000 samples.

\subsection{Results} 
\begin{figure}[htb] 
    \centering
    \includegraphics[width=0.47\textwidth]{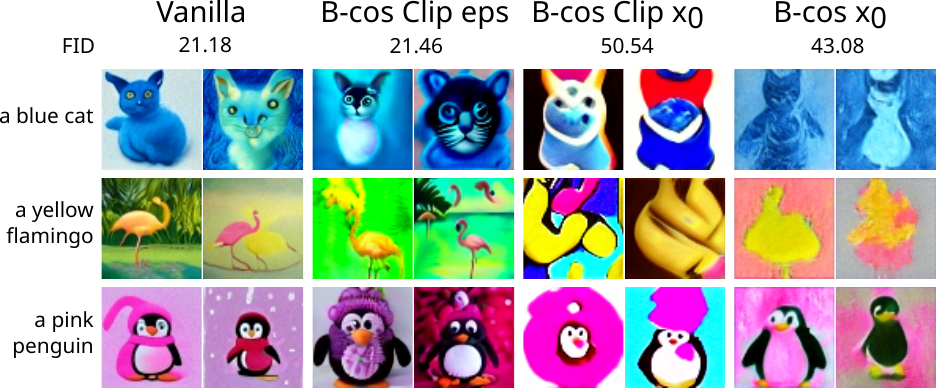}
    \vspace{-8pt}
    \caption{\textbf{Generated samples.} The FID scores and samples generated by the four model configurations using 25 DDIM steps, with two seeds per configuration and prompt.}
    \vspace{-14pt} \label{fig:samples}
\end{figure}

\begin{figure}[htb] 
    \centering
    \vspace{-2pt}
    \includegraphics[width=0.47\textwidth]{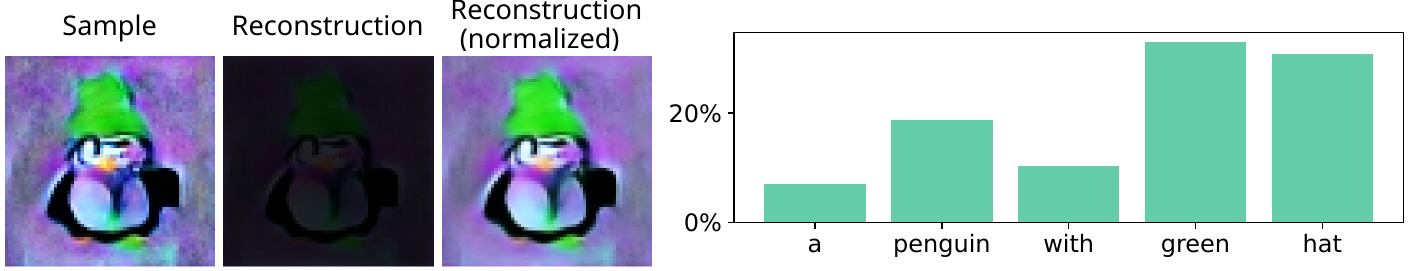}
    \vspace{-8pt}
    \caption{\textbf{Image reconstructions.} The reconstruction, normalized reconstruction and relevance scores of a sample for the prompt \textit{"a penguin with green hat"} with 4 DDIM steps.} 
    \vspace{-2pt}
    \label{fig:recs}
\end{figure}

To evaluate the use of B-cos networks in diffusion models, we assess both the quality of the generated images and the interpretability of the text-to-image relationship. \vspace{-4pt}
\paragraph{Generative Quality}
\Cref{fig:samples} shows that all B-cos networks can successfully generate images even for complex prompts. However, the samples generated by \emph{B-cos Clip eps} are qualitatively better than the ones of the models predicting $x_0$. While the training dataset consisted only of 100'000 images, improving the generative quality is a direction for future research.
\paragraph{Interpretability}
The core idea behind B-cos is that, in the absence of biases, the transformed explanation from \Cref{eq:reconst} should reconstruct the generated image. \Cref{fig:recs} displays reconstructions for a sample. While the raw reconstruction appears noticeably darker than the original image, the normalized version obtained from \Cref{eq:reconst_norm} closely matches the sample, achieving a mean squared error of 0.0045. This strong similarity suggests that the model summary effectively captures the underlying computations. Consequently, even in the presence of biases, the explanation remains faithful and interpretable. Building on this validation, we provide token-level attributions for this sample, which require only a single backward pass. As shown, tokens such as \emph{green}, \emph{hat}, and \emph{penguin} receive higher relevance scores than tokens with weaker semantic meaning like \emph{a} and \emph{with}, indicating that the explanation faithfully reflects the model’s alignment to the prompt.

For a more comprehensive analysis, in \Cref{tab:relevance} we provide the average relevance scores of selected tokens across all prompts of the dataset where they show up without duplicates. Each token occurs over 1,000 times, yet the model assigns notably different relevance scores to them. Importantly, our method consistently attributes higher relevance to semantically meaningful tokens, supporting the global faithfulness of the generated explanations. A broader token-level analysis is presented in \Cref{sec:additional_results}, further demonstrating that tokens with clear semantic content receive higher relevance scores, while less meaningful tokens are attributed lower scores.
\begin{table}[htb] 
    \centering
    \vspace{-4pt}
    \begin{tabular}{cc|cc}
        Token & Relevance & Token & Relevance \\ \hline
        muffins & 30.1\% & to & 4.21\%\\
        flamingo & 27.8\% & a & 3.01\%\\
        banana & 20.6\% & . & 2.88\%\\
        penguin & 17.1\% & you & 1.99\%\\
        goat & 16.3\% & or & 1.87\%\\
        cat & 15.6\% & stock & 1.26\%\\
    \end{tabular}
    \vspace{-7pt}
    \caption{\textbf{Relevance scores.} The average relevance scores of selected tokens indicate that tokens with strong semantic meaning receive higher scores in \emph{B-cos $x_0$}.}
    \vspace{-16pt}
    \label{tab:relevance}
\end{table}

\begin{figure}[htb] 
    \vspace{-4pt}
    \centering
    \includegraphics[width=0.47\textwidth]{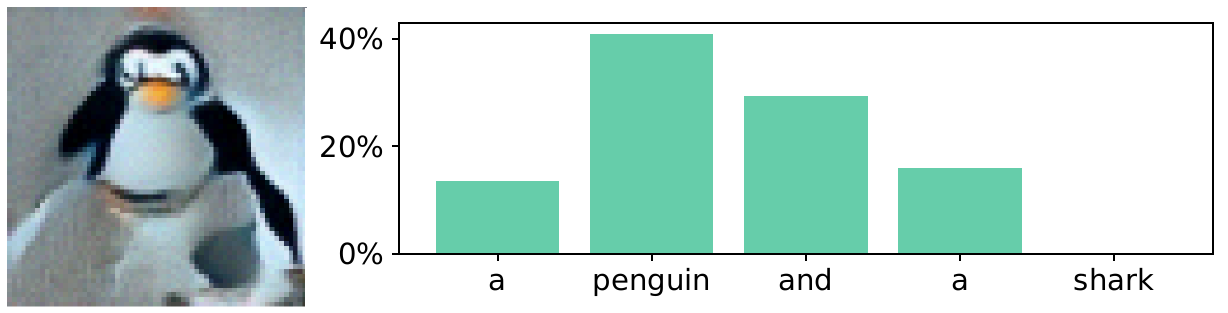}
    \vspace{-9pt}
    \caption{\textbf{Sample and relevance score for an example prompt.} The attribution scores indicate that the image does not fully capture the semantic content of its prompt and thus should be regenerated.}
    \vspace{-14pt}
    \label{fig:attr}
\end{figure}

Having established the faithfulness of our explanations, \cref{fig:attr} presents a failure case in which the model fails to generate a \emph{shark}. The corresponding relevance score (\(0.18\%\)) captures the negligible contribution of the ``shark'' token to the output image. This example demonstrates how the proposed method reliably identifies images that do not align with their prompt, thereby providing a more transparent alternative to CLIP-based scores \cite{hessel2021clipscore}, which depend on pre-trained and poorly understood models for interpretation.

\section{Conclusion}

In this work, we integrate B-cos networks into text-to-image diffusion models. Our results show that B-cos networks can be effectively used in this context. 
While potential biases may arise, we empirically find that model explanations remain largely unaffected. We show that the explanations faithfully capture the alignment between an image and its prompt. Our approach yields token-level attributions that link output pixels to prompt tokens, enabling the detection of missing concepts and guiding the regeneration of incomplete samples. While our approach offers inherent interpretability, it currently leads to reduced generative performance. Developing methods to overcome this trade-off is a promising direction for future research.



\bibliography{example_paper} 
\bibliographystyle{icml2025}

\newpage
\appendix
\onecolumn

\section{Related Work} 
\label{appendix:rel_work}
\textbf{B-cos Networks}

\citet{bohle2022b} introduced B-cos networks with arbitrary neural connections -- including dense and convolutional architectures -- and modules such as activation functions and normalization layers. \citet{bohle2024b} and \citet{tran2023b} independently extend B-cos networks to support self-attention. Though, the former work abandoned the strict requirement of having no bias parameters and find empirically that their models remain highly interpretable despite losing theoretical guarantees. Unlike other works, which exclusively focused on classification models, \citet{arya24bcosification} showed that B-cos networks can also be used for representation learning tasks and developed a technique to interpret the learned representations.

\textbf{Interpretable Diffusion Models}

Despite the rise of diffusion models, few works focus on making them fully interpretable. \citet{li2024self} learn latent concept representations to guide generation and reveal biases, though their goal is mainly content filtering. Others, use cross-attention maps for prompt attribution and image segmentation~\citep{tang2022daam, wang2023diffusion, tian2024diffuse}. While these methods yield intuitive visualizations, the reliability of attention-based explanations remains contested \cite{bibal2022attention}. In contrast, B-cos explanations cover the full computation, extending beyond cross-attention.

\citet{kong2023interpretable} use mutual information to assess word relevance by comparing images generated with and without specific prompt tokens. However, this requires multiple distinct generations and doesn't directly explain a given sample. In contrast, our method analyzes the actual generation process without input perturbations, providing faithful, sample-specific explanations.

\citet{wang2024discrete} propose a framework that leverages token gradients to optimize prompts. However, their approach also relies on a CLIP vision encoder to evaluate prompt adherence. Our method could potentially replace this component or serve as complementary guidance.

\section{Diffusion model}\label{appendix:diffusing}
\newcommand{\N}[1]{\mathcal{N}\left(#1\right)}
Diffusion models are a class of latent variable generative models that learn to reverse a diffusion process that is modeled as a Markov chain. This Markov chain iteratively transforms an image $x_0$ into an image $x_T$ that is indistinguishable from noise by iteratively adding a small amount of noise. In this section, we introduce parameters $\mu$ and $\sigma$ such that the noisy samples $x_T$ resemble samples drawn from the following distribution: $q(x_T \mid x_0) \approx \N{\mu, \sigma^2 I}$. \cite{ho2020denoising} originally propose $\mu=0$ and $\sigma=1$. To use the encoding as described in \cref{subsec:encoding}, we can use $\mu=0.5$. This way, each individual channel of $x_T$ approximates the same distribution. As values outside of the range contribute disproportionally much, we halve the standard deviation to $\sigma=0.5$. However, in this chapter we present the relevant formulas without plugging in the specific values. The forward process can be denoted as follows:

\begin{equation}\label{eq_ftt-1}
q(x_t \mid x_{t-1}) = \N{x_t; \sqrt{\alpha_t}x_{t-1} + (1-\sqrt{\alpha_t})\mu, \sigma^2(1-\alpha_t) I} 
\end{equation}

The parameters $\alpha_t$ are chosen in a suitable way such that $q(x_T \mid x_0) \approx \N{\mu, \sigma^2 I}$ for some desirable mean $\mu$ and variance $\sigma^2$. In our case case, we use the same $\alpha$-parameters as Stable Diffusion, which is a linear schedule with $T=1000$, $\alpha_0 = 0.99915$, and $\alpha_T = 0.988$. Moreover, note that some papers also make use of the parameter $\beta_t = 1-\alpha_t$ in their notation, which we also use for some of the following formulas. 

Instead of iteratively generating a noisy image $x_t$, we can directly sample it from the following distribution where $\bar\alpha_t = \prod_{s=0}^t\alpha_s$:

\begin{equation} \label{eq_ft0}
q(x_t \mid x_0) = \N{x_t; \sqrt{\bar\alpha_t}x_0 + (1-\sqrt{\bar\alpha_t})\mu, \sigma^2(1-\bar\alpha_t) I} 
\end{equation}

By applying the reparameterization trick, one can also rewrite this relation as follows:

\begin{equation} \label{eq_ft0x}
x_t = \sqrt{\bar\alpha_t}x_0 + (1-\sqrt{\bar\alpha_t})\mu + \sigma\sqrt{(1-\bar\alpha_t)}\epsilon_t \quad \text{with}\quad \epsilon_t \sim \N{0,I}
\end{equation}

The reverse process is modeled by the distribution $p_\theta(x_{t-1} \mid x_t)$. Diffusion models employ a neural network that allows approximating this distribution. Instead of predicting $x_{t-1}$ from $x_t$ directly, the neural network typically is used to predict $x_0$, $\epsilon_t$ or another quantity based on those two. Given $x_t$ and (the approximation for) $x_0$, one can then sample $x_{t-1}$ from the following distribution:

\begin{equation} \label{eq_b}
\begin{split}
    &p_\theta(x_{t-1} \mid x_t, x_0) \\ &= \mathcal{N}\left(x_{t-1}; \frac{\sqrt{\alpha_t}(1-\bar\alpha_{t-1})}{1-\bar\alpha_t}x_t + \frac{\sqrt{\bar\alpha_{t-1}}\beta_t}{1-\bar\alpha_t}x_0 \right. +  \left(1 - \frac{\sqrt{\alpha_t}(1-\bar\alpha_{t-1}) + \sqrt{\bar\alpha_{t-1}}\beta_t}{1-\bar\alpha_t}\right)\mu, \quad\left.\frac{\beta_t(1-\bar\alpha_{t-1})\sigma^2}{1-\bar\alpha_t}I\right)
\end{split}
\end{equation}

A training sample for the neural network is thus produced by sampling a timestep $t \sim [1, \dots, T]$, sampling $x_t$ from distribution \ref{eq_ft0} and predicting $\hat x_0$ or some other quantity relating $x_0$ to $x_t$ -- for example the noise $\epsilon_t$. The prediction of $x_{t-1}$ can then be computed using relationship \ref{eq_ft0x} and \ref{eq_b}. Note that these computations may introduce biases. When computing the next step $x_{t-1}$, this resampling introduces biases through $\mu$, the newly introduced noise and previously uncaptured biases in $x_t$. However, note that theses biases are not present during training and hence cannot actively be abused by the model. Moreover, if the neural network predicts $x_0$, the magnitude of the biases is reduced compared to a neural network that predicts $\epsilon_t$. 

To sample a new image using a trained model, given an image $x_T$ consisting of random noise, one can then iteratively apply the above method to obtain an image $x_0$. Originally, \citeauthor{ho2020denoising} proposed $T=1000$, meaning that sampling such an image requires applying the trained neural network a thousand times. This requires a significant amount of computation time. To reduce the number of denoising steps required at inference time, \citeauthor{song2021denoising} introduced DDIM, which allows samples to be generated in an arbitrary number of timesteps independently of the T that was chosen during training. As a consequence of this, images of similar quality can be sampled in 20-50 timesteps. To do so, distribution \ref{eq_b} is replaced by the following distribution at inference time:

\begin{equation}\label{eq_ddim}
\begin{split}
    p_\theta&(x_{t-1} \mid x_t, x_0) \\ &= \mathcal{N}\left(x_{t-1}; \sqrt{\bar\alpha_{t-1}}x_{0}+(1-\sqrt{\bar\alpha_{t-1}})\mu + \sigma\sqrt{1-\bar\alpha_{t-1}-\sigma_t^2}\epsilon_t, \sigma^2\sigma_t^2I\right)
\end{split}
\end{equation}

Note that $x_t$ does not explicitly show up in this formula. However, $x_0$, $x_t$ and $\epsilon_t$ are in a linear relationship as outlined in relationship \ref{eq_ft0x}. Thus, two of them suffice to infer the third one. As $\epsilon_t$ is often the quantity predicted by the neural network, this formulation is closer to the actual implementation.  
DDIM additionally introduces the parameter $\sigma_t$. Notably, $\sigma_t=0$ means that the reverse process is deterministic, which is the only DDIM configuration we use in this work. As a result of this, no additional noise is introduced during the image generation -- so no additional biases are introduced, which enables the model summary to capture more of the computations. 

\section{Model configuration details}\label{sec:model_details}

\begin{figure}[htb] 
    \centering
    \includegraphics[width=0.9\textwidth]{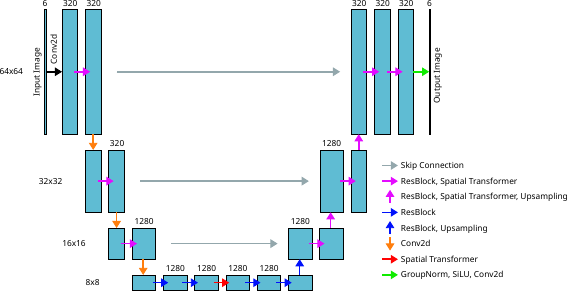}
    \caption{Schematic visualization of our U-Net configuration.}
    \label{fig:unet}
\end{figure}

For the reverse diffusion process, a U-Net~\cite{ronneberger2015u} is used. We built our models on top of Stable Diffusion 2.1 by replacing all modules with their respective B-cos counterparts. The VAE is omitted and therefore the input and output are adjusted to have six instead of four channels. We also reduce the number of residual blocks per downsample step to one instead of two. And we do not double the channels after the first downsampling step. \Cref{fig:unet} shows a visualization of the resulting setup. All linear and convolutional layers are replaced by their B-cos counterparts.

Note that all blocks that include a spatial transformer also receive the encoded prompt as additional input. All our models use pretrained weights of CLIP-ViT-H-14-laion2B-s32B-b79K~\cite{radford2021learning}. For \emph{B-cos $x_0$}, which predicts $x_0$, we only use the embeddings of each token together with the learnt positional embedding and mask all non-prompt tokens. For \emph{B-cos CLIP $x_0$} and \emph{eps}, we use the full pretrained encoder up to the penultimate layer, like Stable Diffusion 2.1. For the attributions in \Cref{sec:additional_results}, we use those final encodings. Thus, the information of some tokens may also be present in other token encodings, such as the encoded \emph{EOS} token, which is trained to represent the full sequence. However, this also allows us to compute such attributions without B-cosifying the CLIP text encoder, which may come with its own challenges and artifacts.

\section{Training details}\label{sec:training}
The models were trained using a GeForce RTX 4090 GPU with the AdamW optimizer set to a learning rate of $2\cdot10^{-6}$. The models were trained for one million steps, taking around three days to complete. A small batch size of 3 was used to accommodate the GPU memory constraints. 

To make training more tractable, we used a subset of the LAION-2B-en-aesthetics dataset \cite{schuhmann2022laion}. We selected five objects -- banana, cat, goat, flamingo, and penguin -- chosen for their high sample availability and ease of human recognition. For each object, we retrieved all captions containing the object’s name in either singular or plural form, excluding cases with leading or trailing alphabetic characters. We then selected the 20'000 image-caption pairs with the highest similarity scores, as provided in the dataset, ensuring strong alignment between images and their descriptions. Images were resized using bilinear interpolation so that the shorter side measured 64 pixels, followed by center cropping. During training, images were randomly flipped horizontally for data augmentation.

\section{Additional Results}\label{sec:additional_results}
\subsection{B-cos Clip eps}

\begin{figure*}[htb]
    \centering
    \begin{subfigure}
      \centering
      \includegraphics[width=.45\linewidth]{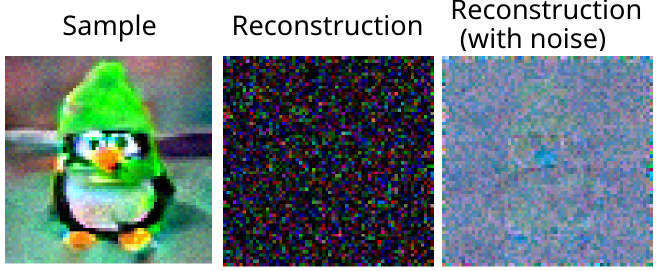}
    \end{subfigure}%
    ~
    \begin{subfigure}
      \centering
      \includegraphics[width=.45\linewidth]{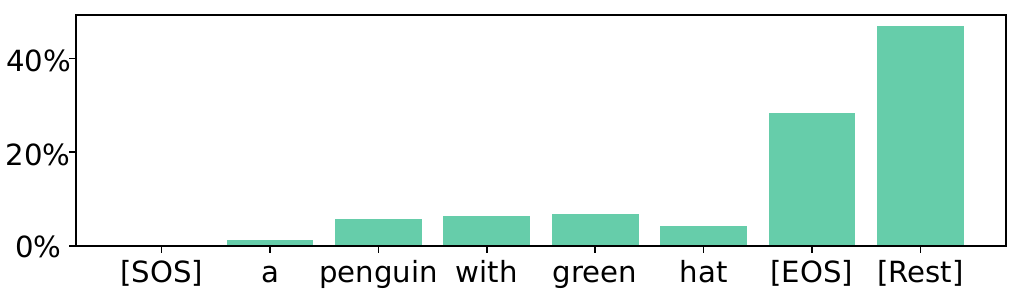}
    \end{subfigure}
    \caption{A sample generated by \emph{B-cos Clip eps} in four timesteps with its reconstruction and the reconstruction when adding the initial noise $x_T$. Computing the normalized reconstruction is not possible as the complementary channels add up to roughly zero. On the right side we display the token attribution where \emph{[REST]} represents the aggregated attributions across all padding tokens.}
    \label{fig:eps}
\end{figure*}

\Cref{fig:samples} shows qualitatively and quantitatively, that the diffusion model sampling quality is significantly better when predicting $\epsilon_t$ instead of $x_0$. In \Cref{fig:eps} we highlight the potential issues of this approach. The reconstruction of the sample resembles noise. 

Normalizing the reconstruction is not really possible, as the first three channels and the last three channels cancel each other out. When predicting $\epsilon_t$ (noise with zero-mean and unit-variance), the corresponding channels should add up to zero, not one. The transformation such that the images are encoded with values between 0 and 1 consists of biases that cannot be captured by the summary. Note that one could predict $\mu + \sigma\epsilon_t$. However, we empirically found that the denoising and sampling quality suffers from such a design choice. 

Nevertheless, the reconstruction itself is in fact not entirely random noise. If we add the noise $x_T$ with some adjustment of the brightness, the shape of the penguin is vaguely visible. This is a consequence of the summary essentially capturing the computations up to the biases -- and one of those biases is the input $x_T$. So the summary does contain some of the relevant computations. However, when inspecting the beak and feet of the penguin, they are orange in the sample but appear bright blue in the reconstruction. So even when compensating for the input bias, there are still other biases that affect the computation beyond some noisiness of the reconstruction. And hence, insights gained when inspecting this model with the help of this dynamic linear summary need to be treated with caution.

For this model, CLIP is used in the same way as in Stable Diffusion. So on the one hand, there are the tokes \emph{SOS} and \emph{EOS} that mark the start and the end of the sequence. On the other hand, the sequence is padded with padding tokens, such that there is a total of 77 tokens. These extra tokens can also contribute to the output. And the attribution shows that the \emph{EOS} token contributes significantly more to the dynamic linear computations than the actual prompt tokens. This is not only the case for this specific example but something we generally observed. We argue that this is not too surprising given that the EOS token was trained to be a representation for the entire sequence. All other token representations were not directly optimized during training. The 70 padding tokens do not contribute much individually but together they still make up roughly 45\% of the dynamic linear computation. So a lot of those computations cannot be attributed to a specific, interesting token. During sampling, the neural network is expected to do two things: it should denoise the input and it should generate something that fits to prompt. The denoising itself is not necessarily dependent on interesting tokens and hence, these attributions could suggest that roughly fifty percent of the computations solely focus on denoising.    

Though, once again, the summary of \emph{B-cos Clip eps} captures significantly less computations compared to the models predicting $x_0$ and the B-cos network does not provide any insights into the interplay of biases, tokens and outputs.

\subsection{B-cos x0}

\begin{figure}[htb] 
    \centering
    \includegraphics[width=0.9\textwidth]{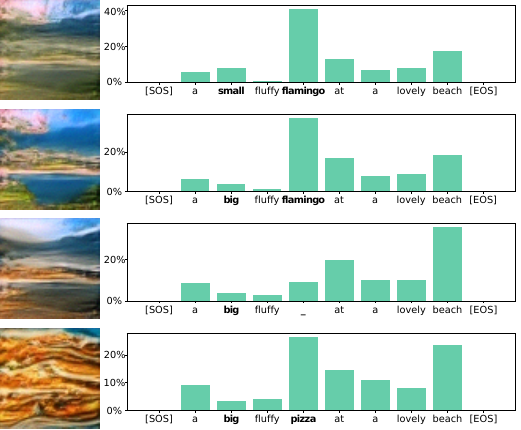}
    \caption{Four samples and their relevance scores generated from the same seed with slightly modified prompts.}
    \label{fig:extras}
\end{figure}

For \Cref{fig:extras} we used the same seed but slightly different prompts to generate four images in 25 timesteps. As the diffusion trajectory is very unstable, the relationship between token relevance and changes in the images is not immediately clear. However, these samples show that replacing the highly relevant token \emph{flamingo} with \emph{pizza} changes the picture completely. 

It also appears that the model does not really make use of the concepts \emph{small} and \emph{big}. Potentially, because it does not understand the objects they describe well enough or because size is relative. Replacing \emph{small} with \emph{big} does lead to minor, but noticeable, colour changes while the structure remains the same. 

The token relevance of \emph{fluffy} is extremely low for many of these images. This suggests that this word is relatively consistently neglected by the model. 

Note that a high token relevance does not mean that the model understands a concept correctly. The model may for example associate the colour purple with the token \emph{flamingo} and hence there is a purple cloud-like structure in the top left corner. 

In \Cref{tab:relevance} we only show a small selection of tokens. \Cref{tab:relevance2} shows the 100 tokens which occur the most often without duplicates in a caption. Note that a word does not necessarily correspond to a single token. For example, the word \emph{flamingos} is split into the two tokens \emph{flamin} and \emph{gos}. The results show that nouns tend to be more relevant than abstract symbols, articles, conjunctions, and similar words. Words such as \emph{stock} or \emph{free} show up in many captions but typically describe the license and not the image itself. Hence, it makes sense that the model does not attribute a lot of relevance to such words. The \emph{SOS}, \emph{EOS} and padding tokens are masked so they cannot contribute to the output. Thus, they also correctly receive zero attribution by the model. With a Pearson correlation coefficient of $-0.145$, there also does not appear to be a simple relationship between the number of occurrences and the relevance of a token. This is consistent with our belief that some words are more important than others.

\begin{table}[htb] 
    \centering
    \begin{tabular}{ccc|ccc}
        Token & \# Occurences & Relevance & Token & \# Occurrences & Relevance \\ \hline
        muffins & 1701 & 30.14\% & [SOS] & 100000 & 0\% \\
        bread & 3690 & 29.55\% & [EOS] & 100000 & 0\% \\
        cheese & 12914 & 29.41\% & stock & 3240 & 1.26\% \\
        flamingo & 14081 & 27.84\% & free & 3435 & 1.64\% \\
        salad & 2433 & 27.07\% & or & 2303 & 1.87\% \\
        pancakes & 2049 & 25.16\% & easy & 1889 & 1.97\% \\
        strawberry & 2674 & 23.28\% & this & 3802 & 2.09\% \\
        cake & 3371 & 22.77\% & summer & 2320 & 2.2\% \\
        vintage & 1674 & 22.48\% & 5 & 4267 & 2.26\% \\
        party & 2808 & 20.81\% & cute & 3453 & 2.34\% \\
        banana & 17897 & 20.63\% & com & 3254 & 2.63\% \\
        girl & 2127 & 20.02\% & one & 1665 & 2.64\% \\
        goats & 3281 & 19.9\% & 4 & 3631 & 2.66\% \\
        painting & 2618 & 19.31\% & / & 2280 & 2.82\% \\
        pink & 4917 & 19.14\% & healthy & 2211 & 2.83\% \\
        flamin & 2905 & 19.1\% & . & 13377 & 2.88\% \\
        penguins & 3155 & 18.39\% & 0 & 5057 & 2.91\% \\
        smoothie & 1778 & 17.29\% & 7 & 2749 & 2.93\% \\
        hat & 2195 & 17.17\% & a & 11837 & 3.01\% \\
        penguin & 10100 & 17.06\% & 6 & 3171 & 3.23\% \\
        goat & 18336 & 16.31\% & is & 3396 & 3.33\% \\
        black & 2025 & 15.67\% & 9 & 3099 & 3.37\% \\
        cat & 17167 & 15.58\% & ... & 2316 & 3.37\% \\
        bananas & 3795 & 15.53\% & vector & 1710 & 3.39\% \\
        dress & 1681 & 15.12\% & ) & 4005 & 3.41\% \\
        cats & 3564 & 14.86\% & are & 1743 & 3.43\% \\
        pattern & 3335 & 14.51\% & 3 & 4252 & 3.44\% \\
        chocolate & 3097 & 14.19\% & ! & 4735 & 3.47\% \\
        blue & 1877 & 13.2\% & an & 1833 & 3.51\% \\
        green & 1717 & 12.44\% & ( & 4670 & 3.62\% \\
        tropical & 3273 & 11.42\% & , & 11534 & 3.69\% \\
        red & 2410 & 9.26\% & 8 & 3184 & 3.78\% \\
        illustration & 2042 & 8.99\% & the & 11579 & 3.85\% \\
        by & 9625 & 8.93\% & for & 8880 & 4.15\% \\
        gos & 2902 & 8.62\% & to & 5429 & 4.21\% \\
        butter & 2151 & 7.83\% & image & 2350 & 4.25\% \\
        christmas & 3309 & 7.71\% & 1 & 7559 & 4.35\% \\
        print & 2257 & 7.45\% & made & 1733 & 4.43\% \\
        birthday & 2202 & 7.31\% & s & 5928 & 4.51\% \\
        at & 3568 & 7.3\% & : & 5114 & 4.54\% \\
        recipe & 3733 & 7.03\% & i & 2451 & 4.66\% \\
        breakfast & 1791 & 6.88\% & of & 10770 & 4.79\% \\
        art & 3350 & 6.51\% & on & 9290 & 4.83\% \\
        photo & 2604 & 6.29\% & \& & 4760 & 4.85\% \\
        and & 26722 & 6.24\% & white & 2992 & 4.85\% \\
        cream & 1782 & 6.04\% & | & 3222 & 5\% \\
        in & 11603 & 5.93\% & from & 3245 & 5.08\% \\
        baby & 2823 & 5.9\% & 2 & 6061 & 5.11\% \\
        with & 24319 & 5.79\% & background & 2466 & 5.15\% \\
        kids & 1796 & 5.69\% & - & 16439 & 5.31\% \\
    \end{tabular}
    \caption{Mean relevance-score of all the 100 most common tokens in the dataset.}
    \label{tab:relevance2}
\end{table}

\end{document}